\documentclass[10pt,twocolumn,letterpaper]{article}

\usepackage{cvpr}
\usepackage{times}
\usepackage{epsfig}
\usepackage{graphicx}
\usepackage{amsmath}
\usepackage{amssymb}
\usepackage{booktabs}
\usepackage{multirow}

\usepackage[pagebackref=true,breaklinks=true,letterpaper=true,colorlinks,bookmarks=false]{hyperref}

\cvprfinalcopy 

\def\cvprPaperID{6049} 

\ifcvprfinal\fi
\begin{document}

\title{Light Field Spatial Super-resolution via Deep Combinatorial Geometry Embedding and Structural Consistency Regularization}


\author{Jing Jin\textsuperscript{\rm 1}
,
Junhui Hou\textsuperscript{\rm 1} \protect\thanks{Corresponding author.} ,
Jie Chen\textsuperscript{\rm 2} ,
Sam Kwong\textsuperscript{\rm 1}\\
\textsuperscript{\rm 1}City University of Hong Kong
\textsuperscript{\rm 2} Hong Kong Baptist University \\
\tt\small jingjin25-c@my.cityu.edu.hk, \{jh.hou,cssamk\}@cityu.edu.hk, chenjie@comp.hkbu.edu.hk
}

\maketitle

\begin{abstract}

Light field (LF) images acquired by hand-held devices usually suffer from low spatial resolution as the limited sampling resources have to be shared with the angular dimension. LF spatial super-resolution (SR) thus becomes an indispensable part of the LF camera processing pipeline. The high-dimensionality characteristic and complex geometrical structure of LF images make the problem more challenging than traditional single-image SR. The performance of existing methods is still limited as they fail to thoroughly explore the coherence among LF views and are insufficient in accurately preserving the parallax structure of the scene. 
In this paper, we propose a novel learning-based LF spatial SR framework, in which each view of an LF image is first individually super-resolved by exploring the complementary information among views 
with combinatorial geometry embedding.
For accurate preservation of the parallax structure among the reconstructed views, a 
regularization network trained over a structure-aware loss function is subsequently appended 
to enforce correct parallax relationships  over the intermediate  estimation. 
Our proposed approach is evaluated over datasets with a large number of testing images including both synthetic and real-world scenes.
Experimental results demonstrate the advantage of our approach over state-of-the-art methods, i.e., our method not only improves the average PSNR by more than 1.0 dB but also preserves more accurate parallax details, at a lower computational cost.

\end{abstract}

\section{Introduction}


4D light field (LF) images differ from conventional 2D images as they record not only intensities but also directions of light rays~\cite{lfapp2017wq4}. The rich information enables a wide range of applications, such as 3D reconstruction~\cite{lfapp2013scene,lfapp2017wq1,lfapp2016wq2,lfapp2017wq3}, refocusing~\cite{lfapp2014refocusing}, and virtual reality~\cite{lfapp2015vr,lfapp2017vryu}.
LF images can be conveniently captured with commercial micro-lens based cameras~\cite{lytro,raytrix} by encoding the 4D LF into a 2D photo detector.
However, due to the limited resolution of the sensor, recorded LF images always suffer from low spatial resolution. Therefore, LF spatial super-resolution (SR) is highly necessary for further applications.

\begin{figure}[t]
\begin{center}
\includegraphics[width=0.85\linewidth]{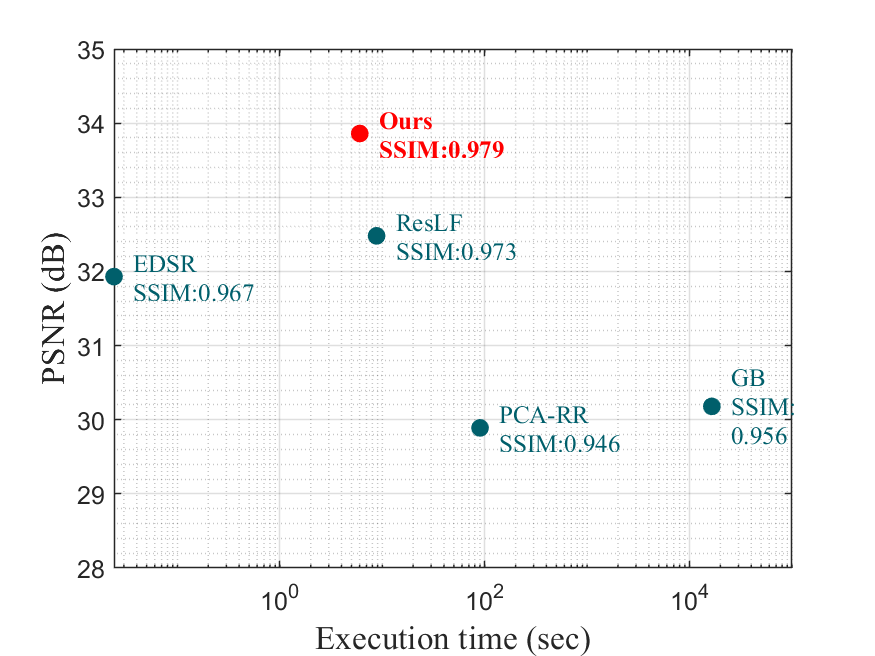}
\end{center} 
\vspace{-0.8em}
  \caption{Comparisons of the running time (in second) and reconstruction quality (PSNR/SSIM) of different methods. The running time is the time for super-resolving an LF image of spatial resolution $94\times135$ and angular resolution $7\times7$ with the scale factor equal to 4. The PSNR/SSIM value refers to the average over 57 LF images in Stanford Lytro Archive (Occlusions) dataset. }
\label{fig:psnr_time}  
\end{figure}

Some traditional methods for LF spatial SR have been proposed 
~\cite{lfssr2014variational,lfssr2017graph,lfssr2012gmm}.
Due to the high dimensionality of LF data, the reconstruction quality of these methods is quite limited.
Recently, some learning-based methods~\cite{yoon2017lfcnn,lfssr2018lfnet,lfssr2019reslf} have been proposed to address the problem of 4D LF spatial SR via data-driven training. 
Although these methods have improved both performance and efficiency, there are two problems unsolved yet.
That is, the complementary information within all views is not fully utilized, and the structural consistency of the reconstruction is not well preserved (see more analyses in Sec.~\ref{sec_motivation}).

In this paper, we propose a learning-based method for LF spatial SR, focusing on addressing the two problems of complete complementary information fusion and LF parallax structure preservation.
As shown in Fig.~\ref{fig:workflow}, our approache consists of two modules, i.e., an All-to-One SR via combinatorial geometry embedding and a structural consistency regularization module.
Specifically, the All-to-One SR module separately super-resolves individual views by learning combinatorial correlations and fusing the complementary information of all views, giving an intermediate super-resolved LF image. 
The regularization module exploits the spatial-angular geometry coherence among the intermediate result, and enforces the structural consistency in the high-resolution space. Extensive experimental results on both real-world and synthetic datasets demonstrate the advantage of our proposed method.
That is, as shown in Fig.~\ref{fig:psnr_time}, our method produces much higher PSNR/SSIM at a higher speed, compared with state-of-the-art methods.

\section{Related Work}


\textbf{Two-plane representation of 4D LFs}. 
The 4D LF is commonly represented using two-plane parameterization.
Each light ray is determined by its intersections with two parallel planes, i.e., a spatial plane $(x,y)$ and a angular plane $(u,v)$.
Let $L(\mathbf{x},\mathbf{u})$ denote a 4D LF image, where $\mathbf{x}=(x,y)$ and $\mathbf{u}=(u,v)$.
A view, denoted as $L_{\mathbf{u}^{\ast}}=L(\mathbf{x},\mathbf{u}^{\ast})$, is a 2D slice of the LF image at a fixed angular position $\mathbf{u}^{\ast}$.
The views with different angular positions capture the 3D scene from slightly different viewpoints.

Under the assumption of Lambertian, projections of the same scene point will have the same intensity at different views. 
This geometry relation leads to a particular \textit{LF parallax structure},
which can be formulated as:
\begin{equation}
\begin{aligned}
    L_{\mathbf{u}}(\mathbf{x}) = L_{\mathbf{u}'}(\mathbf{x}+d(\mathbf{u}' -\mathbf{u} )),
\end{aligned}
\end{equation}
where $d$ is the disparity of the point $L(\mathbf{x},\mathbf{u})$.
The most straightforward representation of the LF parallax structure is epipolar-plane images (EPIs). Specifically, each EPI is the 2D slice of the 4D LF at one fixed spatial and angular position, and consists of straight lines with different slops corresponding to scene points at different depth.

\textbf{LF spatial SR}.
For single image, the inverse problem of SR is always addressed using different image statistics as priors~\cite{sisr2003overview}.
As multiple views are available in LF images, the correlations between them can be used to directly constrain the inverse problem, and the complementary information between them can greatly improve the performance of SR.
Existing methods for LF spatial SR can be classed into two categories: optimization-based and learning-based methods.

Traditional LF spatial SR methods physically model the relations between views based on estimated disparities, and then formulate SR as an optimization problem.
Bishop and Favaro~\cite{lfssr2012bayesian} first estimated the disparity from the LF image, and then used it to 
build an image formation model, which is employed to
formulate a variantional Bayesian framework for SR.
Wanner and Goldluecke~\cite{lfssr2012variational,lfssr2014variational} applied structure tensor on EPIs to estimate disparity maps, which were employed in a variational framework for spatial and angular SR.
Mitra and Veeraraghavan~\cite{lfssr2012gmm} proposed a common framework for LF processing, which models the LF patches using a Gaussian mixture model conditioned on their disparity values.
To avoid the requirement of precise disparity estimation, Rossi and Frossard~\cite{lfssr2017graph} proposed to regularize the problem using a graph-based prior, which explicitly enforces the LF geometric structure.

Learning-based methods exploit the cross-view redundancies and utilize the complementary information between views to learn the mapping from low-resolution to high-resolution views.
Farrugia~\cite{lfssr2017pcarr} constructed a dictionary of examples by 3D patch-volumes extracted from pairs of low-resolution and high-resolution LFs. Then a linear mapping function is learned using Multivariate Ridge Regression between the subspace of these patch-volumes, which is directly applied to super-resolve the low-resolution LF images.
Recent success of CNNs in single image super-resolution (SISR)~\cite{sisr2016srcnn,sisr2017lapsrn,sisr2019survey1} inspired many learning-based methods for LF spatial SR.
Yoon \etal~\cite{lfssr2015lfcnn,yoon2017lfcnn} first proposed to use CNNs to process LF data. They used a network with similar architecture of that in~\cite{sisr2016srcnn} to improve the spatial resolution of neighboring views, which were used to interpolate novel views for angular SR next.
Wang \etal~\cite{lfssr2018lfnet} used a bidirectional recurrent CNN to sequentially model correlations between horizontally or vertically adjacent views. The predictions of horizontal and vertical sub-networks are combined using the stacked generalization technique.
Zhang \etal~\cite{lfssr2019reslf} proposed a residual network to super-resolve the view of LF images. Similar to~\cite{lfdepth2018epinet}, views along four directions are first stacked and fed into different branches to extract sub-pixel correlations. Then the residual information from different branches is integrated for final reconstruction.
However, the performance of side views will be significantly degraded compared with the central view as only few views can be utilized, which will result in undesired inconsistency in the reconstructed LF images.
Additionally, this method requires various models  suitable for views at different angular positions, \eg, 6 models for a $7\times7$ LF image, which makes the practical storage and application harder.
Yeung \etal~\cite{lfssr2018separable} used the alternate spatial-angular convolution to super-resolve all views of the LF at a single forward pass.

\begin{figure}[t]
\begin{center}
\includegraphics[width=\linewidth]{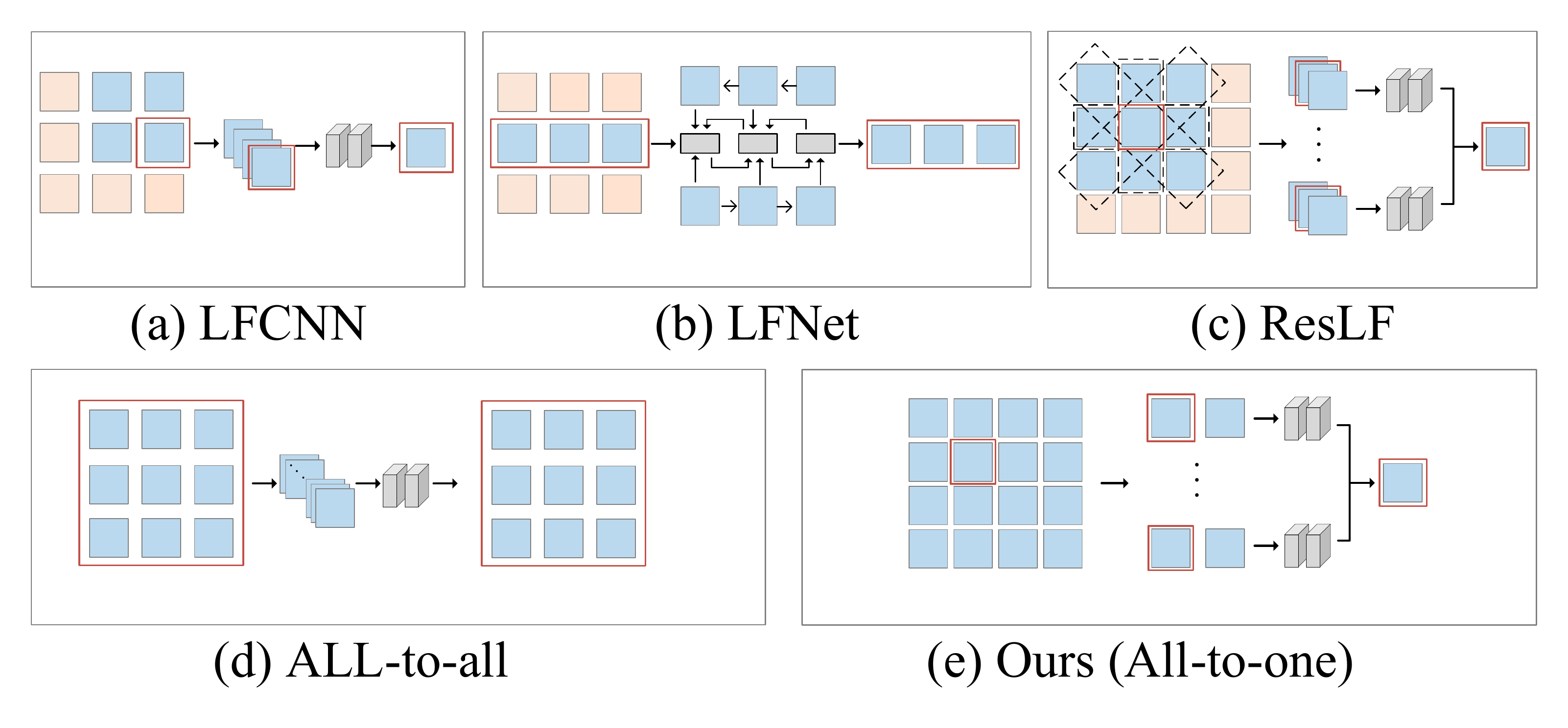}
\end{center} 
\vspace{-1.3em}
  \caption{Illustration of different network architectures for the fusion of view complementary information. (a) LFCNN~\cite{yoon2017lfcnn}, (b) LFNet~\cite{lfssr2018lfnet}, (c) ResLF~\cite{lfssr2019reslf}, (d) an intuitive \textit{All-to-All} fusion (see Sec.~\ref{sec_motivation} ), and (e)  our proposed \textit{All-to-One} fusion. Colored boxes represent images or feature maps of different views. Among them, red-framed boxes are views to be super-resolved, and blue boxes are views whose information is utilized.}
\label{fig:compare} 
\end{figure}


\section{Motivation} 

\label{sec_motivation}
Given a low-resolution LF image, denoted as $L^{lr}\in\mathbb{R}^{H\times W\times M\times N}$,
LF spatial SR aims at reconstructing a super-resolved LF image, 
close to the ground-truth high-resolution LF image $L^{hr}\in\mathbb{R}^{\alpha H\times\alpha W\times M\times N}$, where $H\times W$ is the spatial resolution, $M\times N$ is the angular resolution, and $\alpha$ is the upsampling factor. 
We believe the following two issues are paramount for high-quality LF spatial SR: (1) thorough exploration of the complementary information among views; 
and (2) strict regularization of the \textit{view-level} LF structural parallax. 
In what follows, we will discuss more about these issues, which will shed light on the proposed method.

(1) {\bf Complementary information among views}.
An LF image contains multiple observations of the same scene from slightly varying angles.
Due to occlusion, non-Lambertian reflections, and other factors, the visual information is asymmetric among these observations. In other words, the information absent in one view may be captured by another one,
hence 
all views are potentially helpful for high-quality SR.

Traditional optimization-based methods~\cite{lfssr2012variational,lfssr2014variational,lfssr2017graph,lfssr2012gmm} typically model the relationships among views 
using explicit disparity maps, which is expensive to compute. 
Moreover, inaccurate disparity estimation in occluded or non-Lambertian regions will induce artifacts 
and the correction of such artifacts is beyond the capabilities 
of these optimization-based models.
Instead,
recent learning-based methods, such as LFCNN~\cite{lfssr2015lfcnn}, LFNet~\cite{lfssr2018lfnet} and ResLF~\cite{lfssr2019reslf}, explore the complementary information among views through data-driven training.
Although these methods improve both the reconstruction quality and computational efficiency, the 
complementary information 
among views has not been fully exploited due to the limitation of their view fusion mechanisms.
Fig.~\ref{fig:compare} shows the architectures of different view fusion approaches. LFCNN only uses neighbouring views in a pair or square, while LFNet only takes views in a horizontal and vertical 3D LF. ResLF considers 4D structures by constructing directional stacks, which leaves views not located at the "star" shape un-utilized.

\textit{Remark}. An intuitive way to fully take advantage of the cross-view information is by stacking the images or features of all views, feeding them into a deep network, and predicting the high-frequency details for all views simultaneously.
We refer to this method \textit{All-to-All} in this paper. As illustrated in Fig.~\ref{fig:compare}(d), this is a naive extension of the classical SISR networks~\cite{sisr2016vdsr}.
However, this method 
will compromise unique details that only belong to individual views since it is the average error over all views which is optimized during network training. See the quantitative verification in Sec. \ref{subsec:ablation}. 
To the end, we propose a novel fusion strategy for LF SR, called \textit{All-to-One} SR via \textit{combinatorial geometry embedding}, which super-resolves each individual view by combining the information from all views.

(2) {\bf LF parallax structure}. 
As the most important property of an LF image, the parallax structure should be well preserved after SR. 
Generally, existing methods promote the fidelity of such a structure 
by enforcing corresponding pixels to share similar intensity values.
Specifically, traditional methods employ particular regularization in the optimization formulation, such as the low-rank~\cite{lfssr2014rpca} and graph-based~\cite{lfssr2017graph} regularizer.
Farrugia and Guillemot~\cite{lfssr2019rank} first used optical flow to align all views and then super-resolve them simultaneously via an efficient CNN. However, the disparity between views need to be recovered by warping and inpainting afterwards, which will cause inevitable high-frequency loss.
For most learning-based methods~\cite{lfssr2018lfnet,lfssr2019reslf}, the cross-view correlations are only exploited in the low-resolution space, while the consistency in the high-resolution space is not well modeled. See the quantitative verification in Sec. \ref{subsec:comparisons}.

\textit{Remark}. We address the challenge of LF parallax structure preservation with a subsequent regularization module on the intermediate high-resolution results. Specifically, an additional network is applied to explore the spatial-angular geometry coherence in the high-resolution space, which models the parallax structure implicitly.
Moreover, we use a structure-aware loss function defined on EPIs, which enforces not only view consistency but also models inconsistency on non-Lambertian regions.

\section{The Proposed Method}
\label{sec_method}

\begin{figure*}[t]
\begin{center}
\includegraphics[width=\linewidth]{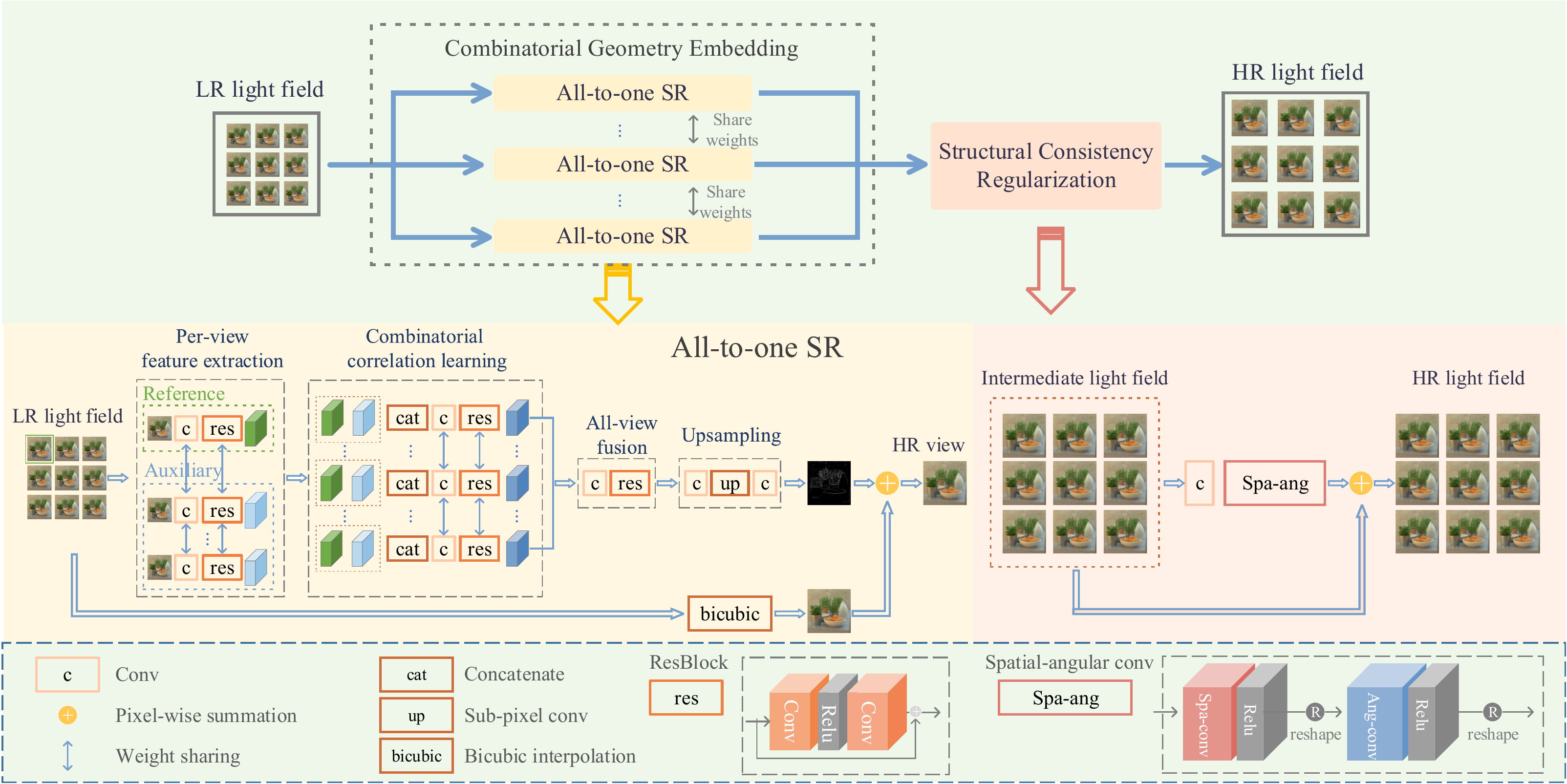}
\end{center} 
\vspace{-1.0em}
  \caption{The flowchart of our proposed approach and illustration of the detailed architecture of the \textit{All-to-One} SR and structural consistency regularization modules. The All-to-One SR module takes full advantage of the complementary information of all views of an LF image by learning their combinatorial correlations with the reference view. At the same time, the unique details of each individual view are also well retained. The structural consistency regularization module recovers the view consistency among the resulting intermediate LF image by exploring the spatial-angular relationships and a structure-aware loss.}
\label{fig:workflow} 
\end{figure*}

As illustrated in Fig. \ref{fig:workflow}, our approach consists of an All-to-One SR module, which  super-resolves each view of an LF image individually by fusing the combinatorial embedding from all other views, and followed by a structural consistency regularization module, which enforces the LF parallax structure in the reconstructed LF image.

\subsection{All-to-One SR via Combinatorial Geometry Embedding}   
Let $L^{lr}_{\mathbf{u}_r}$ denote the reference view to be super-resolved.
The remaining views of an LF image except $L^{lr}_{\mathbf{u}_r}$ are denoted as auxiliary views $\{L^{lr}_{\mathbf{u}_a}\}$.
The All-to-One SR module focuses on extracting the complementary information from auxiliary views to assist the SR of the reference view.
As shown in Fig.~\ref{fig:workflow}, there are four sub-phases involved, i.e., per-view feature extraction, combinatorial correlation learning, all-view fusion and upsampling.

{\bf Per-view feature extraction}.
We first extract deep features, denoted as $F_{\mathbf{u}}^1$, from all views  separately, i.e.,  
\vspace{-0.3em}
\begin{equation}
\begin{aligned}
    F_{\mathbf{u}}^1 = f_1(L_{\mathbf{u}}^{lr}).
\end{aligned}
\end{equation}
Inspired by the excellent performance of residual blocks~\cite{he2016resnet,sisr2016vdsr}, which learn residual mappings by incoporating the self-indentity, we use them for deep feature extraction.
The feature extraction process $f_1(\cdot)$ contains a convolutional layer followed by rectified linear units (ReLU), and $n_1$ residual blocks.
The parameters of $f_1(\cdot)$ are shared across all views.

{\bf Combinatorial correlation learning}.
The geometric correlations between the reference view and auxiliary views vary with their angular positions $\mathbf{u}_r$ and $\mathbf{u}_a$.
To enable our model to be compatible for all views with different $\mathbf{u}_r$ in the LF,
we use the network $f_2(\cdot)$ to learn the correlations between the features of a pair of views $\{F_{\mathbf{u}_1}^1,F_{\mathbf{u}_2}^1\}$, where the angular positions $\mathbf{u}_1$ and $\mathbf{u}_2$ can be arbitrarily selected.
Based on the correlations between $F_{\mathbf{u}_1}^1$  and $F_{\mathbf{u}_2}^1$, $f_2(\cdot)$ is designed to extract information from $F_{\mathbf{u}_2}^1$ and embed it into the features of $F_{\mathbf{u}_1}^1$.
Here, $\mathbf{u}_1$ is set to be the angular position of the reference view, and $\mathbf{u}_2$ can be the position of any auxiliary view. Thus the output can be written as:
\vspace{-0.3em}
\begin{equation}
\begin{aligned}
    F_{\mathbf{u}_r}^{2,\mathbf{u}_a} = f_2(F_{\mathbf{u}_r}^1,F_{\mathbf{u}_a}^1),
\end{aligned}
\end{equation}
where $F_{\mathbf{u}_r}^{2,\mathbf{u}_a}$ is the features of the reference view $L_{\mathbf{u}_r}^{lr}$ incorporated with the information of an auxiliary view $L_{\mathbf{u}_a}^{lr}$.

The network $f_2(\cdot)$ consists of a concatenation operator  to combine the features $F_{\mathbf{u}_r}^1$ and $F_{\mathbf{u}_a}^1$ as inputs, and a convolutional layer followed by $n_2$ residual blocks.
$f_2(\cdot)$'s ability of handling arbitrary pair of views is naturally learned by accepting the reference view and all auxiliary views in each training iteration.



{\bf All-view fusion}.
The output of $f_2(\cdot)$ is a stack of features with embedded geometry information from all auxiliary views.
These features have been trained to align to the reference view, hence they can be fused directly.
The fusion process can be formulated as:
\vspace{-0.4em}
\begin{equation}
\begin{aligned}
    F_{\mathbf{u}_r}^{3} = f_3(F_{\mathbf{u}_{r}}^{2,\mathbf{u}_{a_1}},\cdots,F_{\mathbf{u}_{r}}^{2,\mathbf{u}_{a_m}}),
\end{aligned}
\end{equation}
where $m=MN-1$ is the number of auxiliary views.

Instead of concatenating all features together, we first combine them channel-wise, i.e. combine the feature maps at the same channel across all views. Then, all channel maps are used to extract deeper features.
The network $f_3(\cdot)$ consists of one convolutional layer, $n_{3}$ residual blocks for channel-wise view fusion and $n_{4}$ residual blocks for channel fusion.

{\bf Upsampling}.
We use a similar architecture with residual learning in SISR~\cite{sisr2016vdsr}. 
To reduce the memory consumption and computational complexity, all feature learning and fusion are conducted in low-resolution space. 
The fused features are upsampled using the efficient sub-pixel convolutional layer~\cite{sisr2016espcn}, and a residual map is then reconstructed by a subsequent convolutional layer $f_4(\cdot)$.
The final reconstruction is produced by adding the residual map with the upsampled image: 
\begin{equation}
\begin{aligned}
    L_{\mathbf{u}_r}^{sr} = f_4(U_1(F_{\mathbf{u}_r}^3)) + U_2(L_{\mathbf{u}_r}^{lr}),
\end{aligned}
\end{equation}
where $U_1(\cdot)$ is the sub-pixel convolutional layer and $U_2(\cdot)$ is the bicubic interpolation process.

{\bf Loss function}.
The objective of the All-to-One SR module is to super-resolve the reference view individually  $\widehat{L}_{\mathbf{u}_r}^{sr}$ to approach the ground truth high-resolution image $L_{\mathbf{u}_r}^{hr}$. 
We use the $\ell_1$ error between them to define the loss function:
\vspace{-0.3em}
\begin{equation}
\begin{aligned}
    \ell_{v} =  ||\widehat{L}_{\mathbf{u}_r}^{sr} - L_{\mathbf{u}_r}^{hr} ||_1.
\end{aligned}
\end{equation}

\subsection{Structural Consistency Regularization} 


We apply structural consistency regularization on the intermediate results by the  All-to-One SR module. 
This regularization module employs the efficient alternate spatial-angular convolution to implicitly model cross-view correlations among the intermediate LF images. In addition, a structure-aware loss function defined on EPIs is used to enforce the structural consistency of the final reconstruction. 

{\bf Efficient alternate spatial-angular convolution}.
To regularize the LF parallax structure, an intuitive method is using the 4D or 3D convolution. 
However, 4D or 3D CNNs will result in significant increase of the parameter number and computational complexity.
To improve the efficiency, but still explore the spatial-angular correlations, we adopt the alternate spatial-angular convolution ~\cite{sepfilter2017video,lfssr2018separable,sepfilter2018Yeung}, which handles  
the spatial and angular dimensions in an alternating manner with the 2D convolution.

In our regularization network, we use $n_5$ layers of alternate spatial-angular convolutions.
Specifically, for the intermediate results
$\widehat{L}^{sr} \in \mathbb{R}^{\alpha H\times\alpha W\times M\times N}$, 
we first extract features from each view separately and construct a stack of spatial views, i.e., $F_s \in \mathbb{R}^{\alpha H\times\alpha W\times c\times MN}$, where $c$ is the number of feature maps.
Then we apply 2D spatial convolutions on $F_s$.
The output features are reshaped to the stacks of angular patches, i.e., $F_a \in \mathbb{R}^{M\times N\times c\times \alpha^2HW}$, and then angular convolutions are applied.
Afterwards, the features are reshaped for spatial convolutions, and the previous 'Spatial Conv-Reshape-Angular Conv-Reshape' process repeats $n_5$ times. 

{\bf Structure-aware loss function}.
The objective function is defined as the $\ell_1$ error between the estimated LF image and the ground truth:
\vspace{-0.5em}
\begin{equation}
\begin{aligned}
    \ell_{r} =  || \widehat{L}^{rf} - L^{hr} ||_1,
\end{aligned}
\end{equation}
where $\widehat{L}^{rf}$ is the final reconstruction by the regularization module.

A high-quality LF reconstruction shall have strictly linear patterns on the EPIs. Therefore, to further enhance the parallax consistency, we add additional constraints on the output EPIs. Specifically, we incorporate the EPI gradient loss, which computes the  $\ell_1$ distance between the gradient of EPIs of our final output and the ground-truth LF, for the training of the regularization module. The gradients are computed along both spatial and angular dimensions on both horizontal and vertical EPIs:
\vspace{-0.3em}
\begin{equation}
\begin{aligned}
    \ell_e =  \|\nabla_x \widehat{E}_{y,v} - \nabla_x E_{y,v}\|_1
    +   \|\nabla_u \widehat{E}_{y,v} - \nabla_u E_{y,v}\|_1 \\
    +     \|\nabla_y \widehat{E}_{x,u} - \nabla_y E_{x,u}\|_1 
    +   \|\nabla_v \widehat{E}_{x,u} - \nabla_v E_{x,u}\|_1, 
\end{aligned}
\end{equation}
where $\widehat{E}_{y,v}$ and $\widehat{E}_{x,u}$ denote EPIs of the reconstructed LF images, and
$E_{y,v}$ and $E_{x,u}$ denote EPIs of the ground-truth LF images.

\begin{table}
\label{table:dataset}
\centering
\caption{The datasets used for evaluation.}
\renewcommand{\arraystretch}{1.3}
\resizebox{0.9\linewidth}{!}{
\begin{tabular}{ c| c | c | c }
\toprule[2pt]
~ & Dataset & category & $\#$scenes\\
\midrule[1pt]
\multirow{3}{*}{Real-world}&\multirow{ 2}{*}{Stanford Lytro Archive~\cite{lfdataset2016stanford}} & General & 57 \\
~&~ & Occlusions & 51 \\
\cmidrule{2-4}
~&Kalantari \etal~\cite{lfdataset2016kalantari} & testing & 30\\
\midrule[1pt]
\multirow{2}{*}{Synthetic} & HCI new~\cite{lfdataset2016hci} & testing & 4\\
~&Inria Synthetic~\cite{lfdataset2018inria} & DLFD & 39\\
\bottomrule[2pt]
\end{tabular}
}
\end{table}

\begin{table*}[!t]
\renewcommand{\arraystretch}{1.3}
\caption{Quantitative comparisons (PSNR/SSIM) of different methods on $2\times$ and $4\times$ LF spatial SR. The best results are in bold, and the second best ones are underlined. PSNR/SSIM refers to the average value of all the scenes of a dataset. 
}
\label{table:quan}
\centering
\resizebox{0.94\textwidth}{!}{
\begin{tabular}{c | c|c c c c c c | c }
\toprule[2pt]
~ &  &  Bicubic & PCA-RR~\cite{lfssr2017pcarr} & LFNet~\cite{lfssr2018lfnet} & GB~\cite{lfssr2017graph} & EDSR~\cite{sisr2017edsr} & ResLF~\cite{lfssr2019reslf} &  Ours\\
\midrule[1pt]
Stanford Lytro General~\cite{lfdataset2016stanford} & 2 & 35.93/0.940 & 36.44/0.946 & 37.06/0.952 & 36.84/0.956 & 39.34/0.967 & \underline{40.44/0.973} & \textbf{42.00}/\textbf{0.979} \\
Stanford Lytro Occlusions~\cite{lfdataset2016stanford} & 2 & 35.21/0.939 & 35.56/0.942 & 36.48/0.953 & 36.03/0.947 & 39.44/0.970 & \underline{40.43/0.973} & \textbf{41.92}/\textbf{0.979} \\
Kalantari \etal~\cite{lfdataset2016kalantari} & 2 & 37.51/0.960 & 38.29/0.964 & 38.80/0.969 &  39.33/0.976
 & 41.55/0.980 & \underline{42.95/0.984} & \textbf{44.02}/\textbf{0.987} \\
HCI new~\cite{lfdataset2016hci} &2 &   33.08/0.893 & 32.84/0.883 &33.78/0.904 & 35.27/0.941 & 36.15/0.931 & \underline{36.96/0.946} & \textbf{38.52}/\textbf{0.959} \\
Inria Synthetic~\cite{lfdataset2018inria} &2 &  33.20/0.913 & 32.14/0.885 & 33.90/0.921 & 35.78/0.947 & \underline{37.57}/0.947 & 37.48/\underline{0.953} & \textbf{39.53}/\textbf{0.963} \\
\midrule[1pt]
Stanford Lytro General~\cite{lfdataset2016stanford} &4 &  30.84/0.830 & 31.24/0.841 & 31.30/0.844 & 30.38/0.841 & 33.15/0.882 & \underline{33.68/0.894} & \textbf{34.99}/\textbf{0.917} \\
Stanford Lytro Occlusions~\cite{lfdataset2016stanford} &4 &  29.33/0.794 & 29.89/0.813 & 29.81/0.813 & 30.18/0.855 & 31.93/0.860 & \underline{32.48/0.873} & \textbf{33.86}/\textbf{0.895} \\
Kalantari \etal~\cite{lfdataset2016kalantari} & 4 & 31.63/0.864 & 32.57/0.882 & 32.14/0.879 & 31.86/0.892 & 34.59/0.916 & \underline{35.55/0.930} & \textbf{36.90}/\textbf{0.946} \\
HCI new~\cite{lfdataset2016hci} &4 &  28.93/0.760 & 29.29/0.776 & 29.31/0.773 & 28.98/0.789 & 31.12/0.819 & \underline{31.38/0.838} & \textbf{32.27}/\textbf{0.859} \\
Inria Synthetic~\cite{lfdataset2018inria} & 4 & 28.45/0.795 & 28.71/0.792 & 28.91/0.809 & 29.12/0.836 & \underline{31.68}/0.865 & 31.62/\underline{0.872} & \textbf{32.72}/\textbf{0.890}\\
\bottomrule[2pt]
\end{tabular}
} 
\end{table*}

\begin{figure}[t]
\begin{center}
\includegraphics[width=0.8\linewidth]{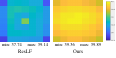}
\end{center}
\vspace{-1.5em}
\caption{Comparison of the PSNR of the individual reconstructed view in \textit{Bedroom}. The color of each grid represents the PSNR value.}
\label{fig:view_psnr} 
\end{figure}

\begin{figure*}[t]
\begin{center}
\includegraphics[width=0.94\linewidth]{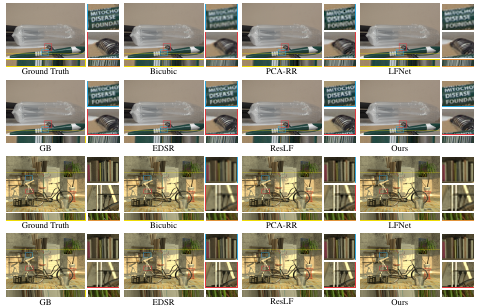}
\end{center}
\vspace{-1em}
  \caption{Visual comparisons of different methods on $2\times$ reconstruction. The predicted central views, the zoom-in of the framed patches, the EPIs at the colored lines, and the zoom-in of the EPI framed patches in EPI are provided. Zoom in the figure for better viewing.
  }
\label{fig:visual_x2} 
\end{figure*}

\begin{figure*}
\begin{center}
\includegraphics[width=0.94\linewidth]{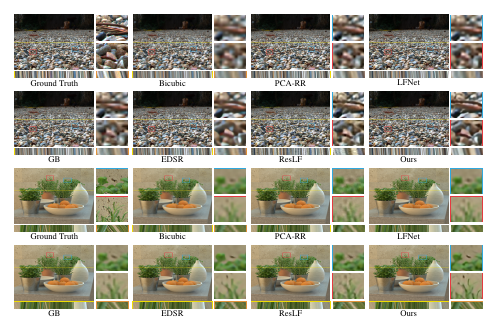}
\end{center}
\vspace{-1.5em}
  \caption{Visual comparisons of different methods on $4\times$ reconstruction.  The predicted central views, the zoom-in of the framed patches, the EPIs at the colored lines, and the zoom-in of the EPI framed patches in EPI are provided. Zoom in the figure for better viewing.}
\label{fig:visual_x4}
\end{figure*}

\subsection{Implementation and Training Details}
{\bf Training strategy}.
To make the All-to-One SR module compatible for all different angular positions, we first trained it independently from the regularization network.
During training, a training sample of an LF image was fed into the network, while a view at random angular position was selected as the reference view.
After the All-to-One SR network training was complete, we fixed its parameters and used them to generate the intermediate inputs for the training of the subsequent structural consistency regularization module. The code is available at https://github.com/jingjin25/LFSSR-ATO. 

{\bf Parameter setting}.
In our network, each convolutional layer has 64 filters with kernel size $3\times3$, and zero-padding was applied to keep the spatial resolution unchanged.
In the per-view SR module, we set $n_1=5$, $n_2=2$, $n_{3}=2$ and $n_{4}=3$ for the number of residual blocks.
For structural consistency regularization, we used $n_5=3$ alternate convolutional layers.

During training, we used LF images with angular resolution of $7\times7$, and randomly cropped LF patches with spatial size $64\times64$.
The batch size was set to 1.
Adam optimizer~\cite{kingma2014adam} with $\beta_1=0.9$ and $\beta_2=0.999$ was used.
The learning rate was initially set to $1e^{-4}$ and decreased by a factor of 0.5 every 250 epochs.

{\bf Datasets}.
Both synthetic and real-world LF datasets were used for training (180 LF images in total, which includes 160 images from Stanford Lytro LF Archive~\cite{lfdataset2016stanford} and Kalantari \eg~\cite{lfdataset2016kalantari}, and 20 synthetic images from HCI~\cite{lfdataset2016hci}). 
We used the bicubic down-sampling method to generate low-resolution images.


\section{Experimental Results}

4 LF datasets containing totally 138 real-world scenes and 43 synthetic scenes were used for evaluation. 
Details of the datasets and categories were listed in Table~\ref{table:dataset}.
Only Y channel was used for training and testing, while Cb and Cr channels were upsampled using bicubic interpolation when generating visual results.

\subsection{Comparison with State-of-the-art Methods}
\label{subsec:comparisons}
We compared with 4 state-of-the-art LF SR methods, including 1 optimization-based method, i.e., GB~\cite{lfssr2017graph}, 3 learning-based methods, i.e., PCA-RR~\cite{lfssr2017pcarr}, LFNet~\cite{lfssr2018lfnet}, and ResLF~\cite{lfssr2019reslf}, 
and 1 advanced SISR method EDSR~\cite{sisr2017edsr}.
Bicubic interpolation was evaluated as baselines.

\textbf{Quantitative comparisons of reconstruction quality}. PSNR and SSIM are used as the quantitative indicators for comparisons, and the average PSNR/SSIM over different testing datasets were listed in 
Table~\ref{table:quan}.
It can be seen that 
our method outperforms the second best method, i.e. ResLF, by around 1 - 2 dB on both $2\times$ and $4\times$ SR.

We also compared the PSNR of individual views between
ResLF~\cite{lfssr2019reslf} and ours, as shown in Figure ~\ref{fig:view_psnr}. It can be observed that the gap between the central and corner views of our method is much smaller than that of ResLF. 
The significant degradation of the corner views in ResLF
is caused by decreasing the number of views used for constructing directional stacks. Our method avoids this problem by utilizing the information of all views. Therefore, the performance degradation is greatly alleviated.

\textbf{Qualitative comparisons}. We also provided visual comparisons of different methods, as shown in Fig.~\ref{fig:visual_x2} for $2\times$ SR and Fig.~\ref{fig:visual_x4} for $4\times$ SR.
It can be observed that most high-frequency details are lost in the reconstruction results of some methods, including PCA-RR, LFNet and GB.
Although EDSR and ResLF could generate better results, some extent of blurring effects occurs in texture regions, such as the characters in the pen, the branches on the ground and the digits on the clock.
In contrast, our method can produce SR results with sharper textures closer to the ground truth ones, which demonstrates higher reconstruction quality. 

\textbf{Comparisons of the LF parallax structure}.
As we discussed in Sec.~\ref{sec_motivation}, the straight lines in EPIs provide direct representation for the LF parallax structure.
To compare the ability to preserve the LF parallax structure, the EPIs constructed from the reconstructions of different methods were depicted in Fig.~\ref{fig:visual_x2} and Fig.~\ref{fig:visual_x4}.
It can be seen that the EPIs from our methods show  clearer and more consistent straight lines compared with those from other methods.

\begin{figure}[t]
\begin{center}
\includegraphics[width=0.45\linewidth]{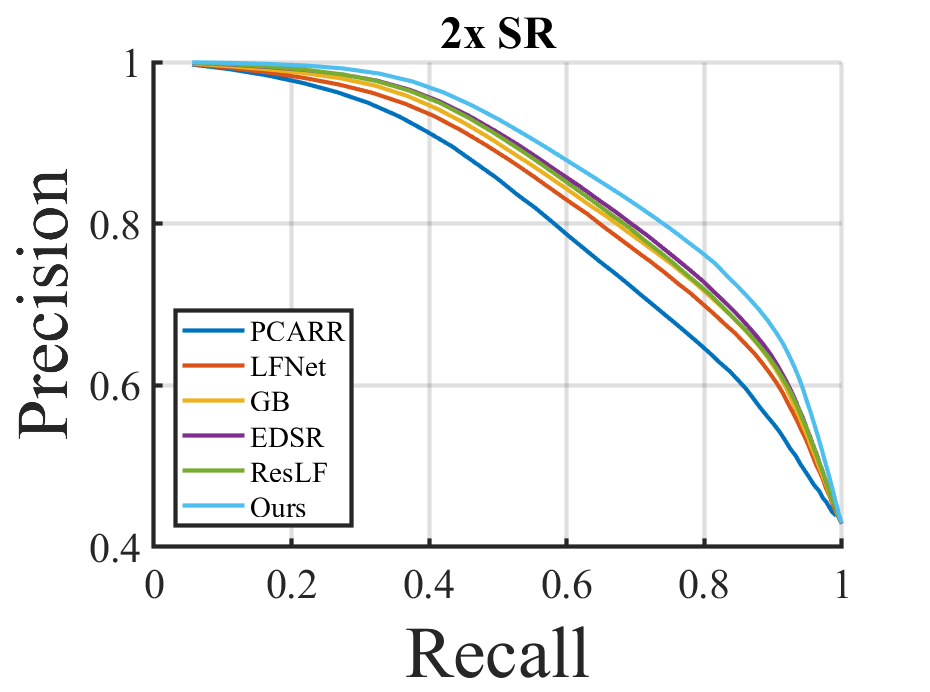}
\includegraphics[width=0.45\linewidth]{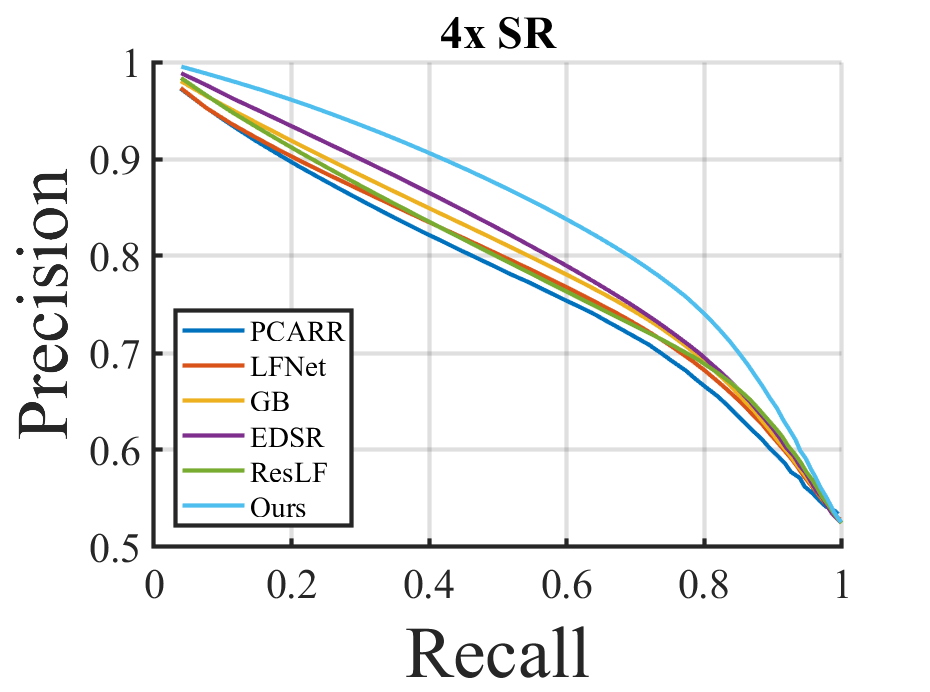}
\end{center} 
\vspace{-1em}
  \caption{Quantitative comparisons of the LF parallax structure of the reconstruction results of different methods via LF edge parallax PR curves. The closer to the top-right corner the lines are, the better the performance is.
  }
\label{fig:prcurve} 
\end{figure}

Moreover, to quantitatively compare the structural consistency, we computed the light filed edge parallax precision-recall (PR) curves~\cite{lfapp2018denoising}, and Fig.~\ref{fig:prcurve} shows the results.
The PR curves of the reconstructions by our method are closer to the top-right corner, which demonstrates the advantage of our method on structural consistency.

\begin{table}[!t]
\renewcommand{\arraystretch}{1.3}
\caption{Investigation of the effectiveness of the structural consistency regularization. The comparisons of the reconstruction quality before and after the regularization are listed. The top two rows show the average PSNR/SSIM over three datasets, and the rest rows show the comparisons on several LF images. }
\label{table:refine}
\centering
\resizebox{\linewidth}{!}{
\begin{tabular}{c | c c }
\toprule[2pt]
 ~ & w/o regularization & w/ regularization\\
\midrule[1pt]
Stanford Lytro Occlusions & 41.62/0.978 & 41.92/0.979 \\
HCI new &  38.24/0.956 & 38.52/0.959 \\
\midrule[1pt]
Occlusion$\_$43$\_$eslf & 33.87/0.986 & 34.53/0.987 \\
Occlusions$\_$51$\_$eslf & 42.14/0.987 & 42.60/0.988 \\
Antiques$\_$dense & 46.31/0.987 & 46.83/0.988 \\
Blue$\_$room$\_$dense & 40.19/0.978 & 40.66/0.979 \\
Coffee$\_$time$\_$dense & 35.09/0.977 & 35.65/0.979 \\
Rooster$\_$clock$\_$dense & 42.45/0.979 &  42.90/0.981 \\
Cars & 38.89/0.986 & 39.33/0.987 \\
IMG$\_$1554$\_$eslf & 37.02/0.989 & 37.46/0.991 \\
\bottomrule[2pt]
\end{tabular}
}  
\end{table}


\begin{table}
\renewcommand{\arraystretch}{1.1}
\caption{Comparisons of running time (in second) of different methods.}
\label{table:time}
\centering
\resizebox{\linewidth}{!}{
\begin{tabular}{c | c c c c c | c }
\toprule[2pt]
~ & \multirow{2}{*}{Bicubic}  & PCA-RR   & GB  & EDSR & ResLF  & \multirow{2}{*}{Ours}\\
~ & ~ & \cite{lfssr2017pcarr} & \cite{lfssr2017graph} &  \cite{sisr2017edsr}  &  \cite{lfssr2019reslf} & ~ \\
\midrule[1pt]
$2\times$ & 1.45 & 91.00 & 17210.00 & 0.025 & 8.98 & 23.35 \\
\midrule[1pt]
$4\times$ &  1.43 & 89.98 & 16526.00 & 0.024 & 8.79 & 7.43\\
\bottomrule[2pt]
\end{tabular}
} 
\end{table}
\textbf{Efficiency comparisons}.
We compared the running time of different methods, and Table~\ref{table:time} lists the results of $4\times$ reconstruction.
Among them, learning-based methods were accelerated by a GeForce RTX 2080 Ti GPU.
SISR methods, i.e., EDSR and bicubic, are faster than other compared methods, as all views can be processed in parallel.
Although our method and ResLF are slightly slower than these SISR methods, much higher reconstruction quality is provided. 

\subsection{Ablation Study}
\label{subsec:ablation}

\textbf{\textit{All-to-One} vs. \textit{All-to-All}}.
We compared the reconstruction quality of our proposed \textit{All-to-One} fusion strategy with the intuitive \textit{All-to-All} one, which simultaneously super-resolves all views by stacking the images or features of all views as inputs to a deep network.
These networks were set to contain the same number of parameters for fair comparisons.
Table~\ref{table:sr} lists the results, where it can be seen that our \textit{All-to-One} improves the PSNR by more than 0.6 dB on real-world data and 1.0 db on synthetic data, respectively, validating its effectiveness and advantage.

\begin{table}[!t]
\renewcommand{\arraystretch}{1.5}
\caption{Comparisons of the intuitive \textit{All-to-All} fusion strategy and our \textit{All-to-One}.}
\label{table:sr}
\centering
\resizebox{\linewidth}{!}{
\begin{tabular}{c | c c c}
\toprule[2pt] 
 & \textit{All-to-All} (image) & \textit{All-to-All} (feature) & Ours \textit{All-to-One}\\
\midrule[1pt]
General & 41.25/0.977 &  41.25/0.977 & \textbf{41.81/0.979}     \\
 Kalantari & 43.18/0.985  &  43.13/0.985 & \textbf{43.79/0.987}  \\
HCI new & 37.12/0.946 & 37.04/0.946  & \textbf{38.24/0.956}\\
\toprule[2pt] 
\end{tabular}
} 
\end{table}

\textbf{Effectiveness of the structural consistency regularization}.
We compared the reconstruction quality of the intermediate (before regularization) and final results (after regularization), as listed in Table~\ref{table:refine}.
It can be observed that around 0.2-0.3 dB improvement is achieved on average over various datasets.
For certain scenes, the contribution of the regularization is more obvious, such as 'Occlusion$\_$43$\_$eslf' and 'Antiques$\_$dense', which obtain more than 0.5dB improvement by the regularization.

\section{Conclusion and Future Work}
We have presented a learning-based method for LF spatial SR. 
We focused on addressing  two crucial  problems, which we believe are paramount for high-quality LF spatial SR, i.e., how to fully take advantage of the complementary information among views, and how to preserve the LF parallax structure in the reconstruction. By modeling them with two sub-networks, i.e., All-to-One SR via combinatorial geometry embedding and structural consistency regularization, our method efficiently generates super-resolved LF images with higher PSNR/SSIM and better LF structure, compared with the state-of-the-art methods.

In our future work, other loss functions, such as the adversarial loss and the perceptual loss which have proven to promote realistic textures in SISR, and their extension to high-dimensional data can be exploited in LF processing.

\clearpage

{\small
\bibliographystyle{ieee_fullname}
\bibliography{reference}
}

\end{document}